\pdfoutput=1
\documentclass[10pt,twocolumn,letterpaper]{article}

\usepackage{cvpr}
\usepackage{graphicx}
\usepackage{amsmath}
\usepackage{amssymb}
\usepackage{algorithm}
\usepackage{algorithmic}
\usepackage{booktabs}
\usepackage{array}
\usepackage{tabularx}
\usepackage{multirow}
\usepackage{hhline}
\usepackage{caption}
\usepackage{subcaption}
\usepackage{tcolorbox}
\tcbuselibrary{skins,breakable}
\usepackage{xspace}

\makeatletter
\DeclareRobustCommand\onedot{\futurelet\@let@token\@onedot}
\def\@onedot{\ifx\@let@token.\else.\null\fi\xspace}

\makeatother

\definecolor{cvprblue}{rgb}{0.21,0.49,0.74}
\usepackage[pagebackref,breaklinks,colorlinks,allcolors=cvprblue]{hyperref}

\def\paperID{*****}
\def\confName{CVPR}
\def\confYear{2026}

\title{\textit{\textbf{Iron}}: Intent-Aligned and Retrospective Dual Learning Framework for Enhancing Generalist Virtual Agents}

\author{
Jiahe Ying$^{1}$ \quad Wendong Bu$^{2}$ \quad Kaihang Pan$^{2}$ \quad Bingchen Miao$^{2}$ \quad Siyu Chen$^{2}$\\
Wen Wang$^{2}$ \quad Xueming Jiang$^{2}$ \quad Juncheng Li$^{2,*}$ \quad Siliang Tang$^{2,*}$\\[3pt]
$^{1}$Fudan University \qquad $^{2}$Zhejiang University\\
$^{*}$Corresponding authors
}

\begin{document}
\maketitle

\begin{abstract}
Achieving virtual agents capable of automating tasks across diverse digital environments remains a pivotal challenge in Embodied AI. While Multimodal Large Language Models (MLLMs) offer enhanced visual perception and reasoning, their agentic deployment faces three challenges: costly data annotation, imprecise action-intent alignment, and inefficient exploration from discarded failed trajectories. To address these, we introduce Iron, an intent-aligned, self-improved, and annotation-efficient framework for training GUI agents. Iron employs a novel dual learning strategy that utilizes a stepwise cycle-consistent (SCC) reward to achieve fine-grained alignment between low-level actions and high-level intents, thereby improving instruction grounding and intent understanding. Concurrently, Iron introduces a hindsight reproduction mechanism to repurpose failed trajectories for training, improving both learning efficiency and task diversity. Extensive experiments demonstrate that Iron-trained generalist agents consistently improve performance on cross-environment and cross-device tasks, outperforming models trained with three times more data. Iron also achieves a substantial 25.06\% relative improvement on unseen web tasks, with further gains observed on inherently complex tasks, demonstrating the feasibility of building more capable virtual agents.
\end{abstract}

\section{Introduction}
\label{sec:intro}

\begin{figure}[t!]
    \centering

    \begin{subfigure}[b]{0.3\columnwidth}
        \centering
        \includegraphics[width=\textwidth]{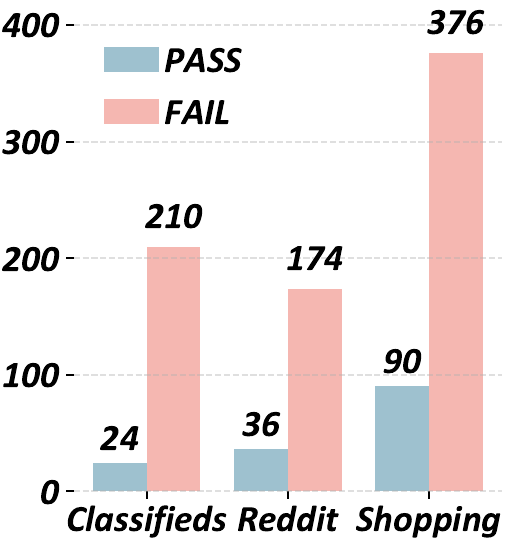}
        \caption{}
        \label{fig:a}
    \end{subfigure}
    \hfill 
    \begin{subfigure}[b]{0.68\columnwidth}
        \centering
        \includegraphics[width=\textwidth]{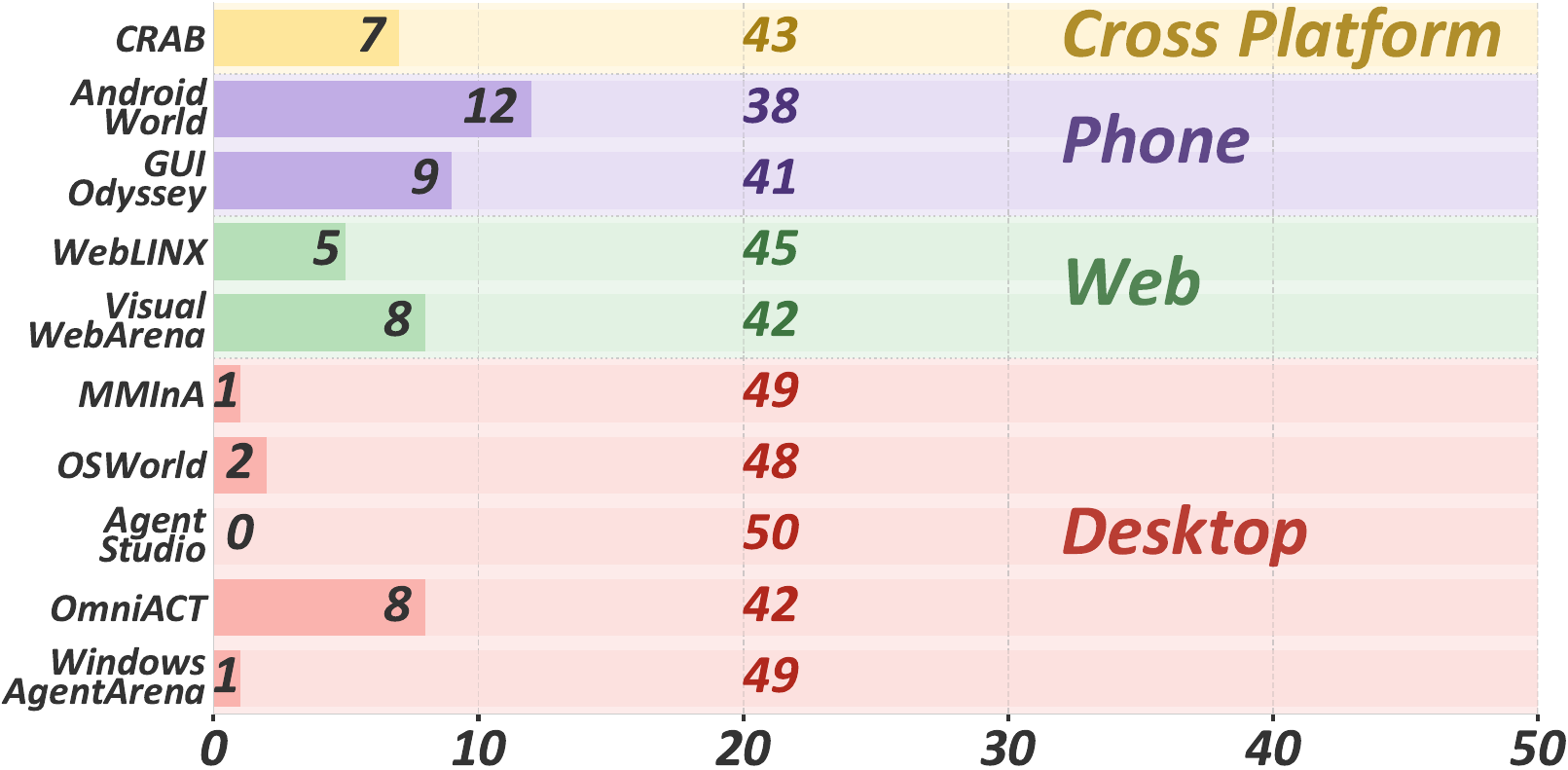}
        \caption{}
        \label{fig:b}
    \end{subfigure} 
    
    
    \begin{subfigure}[b]{\columnwidth}
        \centering
        \includegraphics[width=\textwidth]{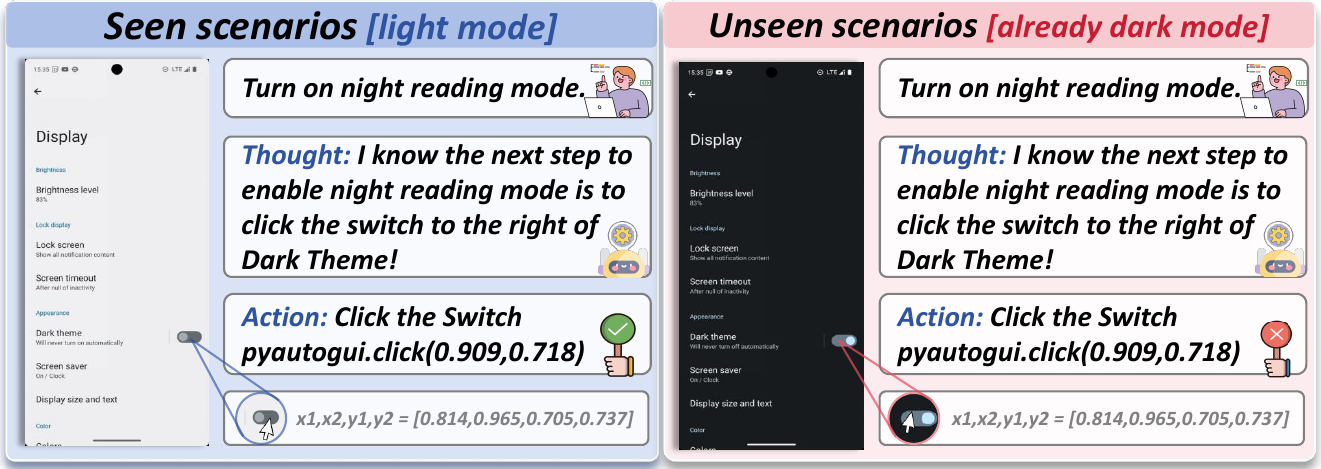} 
        \caption{} 
        \label{fig:c}
    \end{subfigure}
    
    \caption{(a) Scarcity of successful trajectories in GPT-4V's annotation of VisualWebArena. (b) In 50 sampled GPT-4o trajectories, failures (right bar) vastly outnumber successes (left bar). (c) Comparison between the `Mechanical Imitation' of traditional agents and an \textit{Iron-trained} agent in both similar and unseen scenarios.}
    \label{fig:combined}
\end{figure}

Building virtual agents to automate tasks in diverse digital environments has been a long-standing goal. Compared to traditional LLM-based agents~\citep{agentq, webarena, mind2web, RAP}, which are strictly limited to textual data (such as HTML or XML) and often struggle to accurately understand UI visuals, recent advances in MLLMs~\citep{Qwen-VL, GPT4V, llama} have significantly expanded their capabilities in perceiving, reasoning, and acting on visual information. This makes MLLMs inherently more suitable for assuming the roles of general virtual agents.


    
        
        
        
        
    
    
    

Despite these advancements, significant challenges persist in three key areas:
\textbf{1) High costs for labeling superior, diverse sequential decision data:} The complexity of the digital environment poses a significant challenge for human efforts. While MLLMs have been proposed for automated labeling~\citep{koh2024visualwebarena}, their deficiencies in multi-image understanding and complex contextual reasoning render them unreliable as label sources, which results in very limited usable data (see Figure~\ref{fig:a}). Moreover, insufficient reward signals further hinder virtual agents' automated annotation. \textbf{2) Inefficient exploration caused by numerous wasted failed trajectories:} To tackle very limited data, some methods integrate reward models with Monte Carlo Tree Search (MCTS) for autonomous data collection~\citep{agentq, AGENTR, ExAct}. However, their reliance on sampling only successful trajectories creates a critical inefficiency by ignoring the vast number of failed ones. Our own experiments validate this: when sampling 50 trajectories with GPT-4o~\cite{GPT4o}, we found that failures vastly outnumber successes across diverse environments (Figure~\ref{fig:b}), demonstrating that even advanced MLLMs predominantly fail during exploration. This issue is magnified in challenging tasks where successful trajectories may be entirely absent~\citep{agentstudio}. This leads to a significant waste of resources and impedes performance on complex tasks because of a lack of diverse examples and reference demonstrations. \textbf{3) Lack of fine-grained alignment between low-level actions and underlying intent:} Most current MLLM-based agents merely fit the existing training trajectories in a ``mechanical imitation'' way~\citep{cai2024groot2weaklysupervisedmultimodal}. That is, although given a human instruction and a complete trajectory, the agent does not understand what sub-goal a particular trajectory segment accomplishes and how it contributes to the overall objective, thus leading to poor performance in unseen scenarios, as illustrated in the example shown in Figure~\ref{fig:c}.  Hence, the agent needs fine-grained, stepwise low-level sequence intent understanding. 

To address the above challenges, we propose \textit{\textbf{Iron}}, an \underline{I}ntent-aligned and \underline{R}etr\underline{o}spective Self-trai\underline{n}ing Framework. Without requiring manual labor, it employs a dual learning strategy~\cite{xia2016duallearningmachinetranslation,chen2024dualreflectenhancinglargelanguage} to achieve fine-grained alignment between low-level actions and underlying intents, simultaneously utilizing failure data through hindsight reproduction and improving the agent iteratively in a self-training paradigm. Specifically, \textbf{we explicitly formulate two complementary tasks to achieve the dual learning strategy}: 

\textit{(i) Instruction Grounding}: translate instruction $I$ into a sequence of executable operations $A$. 

\textit{(ii) Intent Understanding}~\citep{inferhumansintentionsfollowing, handmethat}: infer the intent $U$ in the given sequence of executable operations $A$ in \textit{(i)}. 

\begin{figure*}[t] 
    \centering     
    \includegraphics[width=\textwidth]{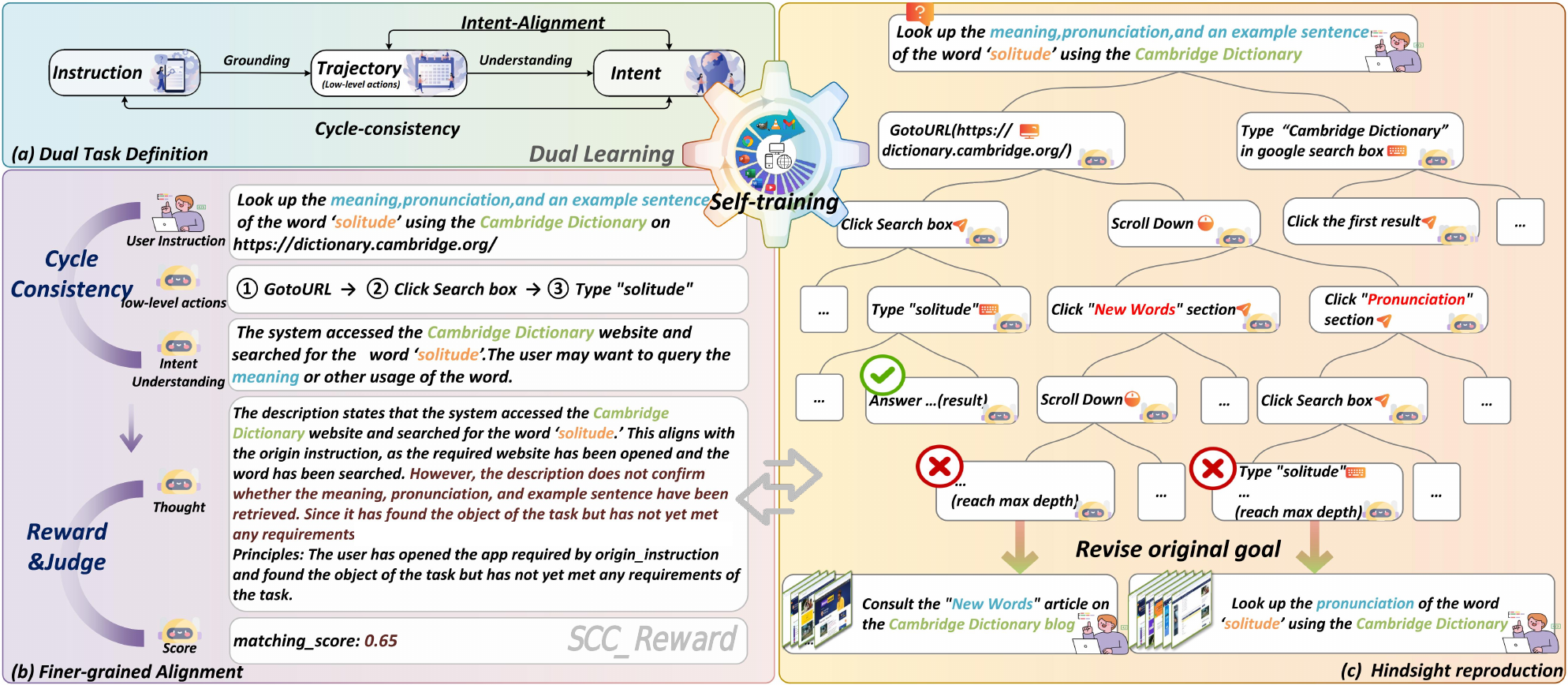} 
    \caption{\textbf{High-level overview of our \textit{Iron} framework.} (a) We first define two complementary and mutually promoting tasks from the Dual Learning strategy.\label{fig2a} (b) Then we achieve fine-grained Intent-Action alignment by introducing a stepwise cycle-consistent reward mechanism. (c) Thirdly, in the process of MCTS, we used hindsight reproduction to transform failed trajectories into high-diversity, actionable training signals fully. Finally, \textit{Iron} continuously optimizes the model through self-training. The illustration features an expansion of two child nodes for exemplary purposes.}\label{fig:overview1_full_page}  
    
\end{figure*}

As shown in Figure~\ref{fig2a}a, the two tasks establish a dual cycle, mutually reinforcing each other to yield synergistic benefits in the training stage (detailed in subsequent sections). We then capitalize on the inherent cycle-consistency between these tasks as a core mechanism to propose the stepwise cycle-consistent (SCC) reward. This design provides a per-step self-assessment signal, ensuring a deep, fine-grained action-intent alignment, which facilitates autonomous learning without relying on external labeling.

Additionally, building upon the fundamental realization that every trajectory culminates in a valid, concrete outcome, regardless of whether it achieves the intended goal, we implement a hindsight reproduction mechanism within MCTS to effectively leverage failure experiences ( Figure~\ref{fig:overview1_full_page}c). By utilizing the inferred intent ($U$), a failed trajectory is thus repurposed into a ``successful trajectory with a revised goal". For further optimization, we designed five key metrics (\textit{Repeatability, Logicality, Ineffectiveness, Exploratory,  Invalidity}) to filter useless converted  trajectories. This innovative mechanism boosts both MCTS utilization and exploration efficacy, ultimately establishing a high-quality, scalable data collection paradigm that substantially augments data volume and task diversity.

Driven by the aforementioned innovations, we conducted extensive experiments leveraging Qwen3-VL-8B~\citep{qwen3vl}, OS-Atlas-Base-7B~\citep{osatlas}, and InternVL2.5-4B~\citep{internvl25} as backbones. \textit{Iron} delivers superior cross-platform results, consistently outperforming the Step-DPO variant and the baseline trained with triple the SFT data. Notably, on the stronger Qwen3-VL-8B backbone, \textit{Iron} achieves 41.96\% on OSWorld and 39.15\% on AndroidWorld, while maintaining over \textbf{25\%} relative improvement in unseen web scenarios. This robust performance is particularly evident on complex tasks. Our main contributions are threefold:


\begin{itemize}
    \item We introduce \textit{Iron}, a novel self-training framework that allows agents to learn from both successes and failures via action-intent alignment and hindsight reproduction.

    \item As a model-agnostic framework, \textit{Iron} can be easily applied to various agent architectures, providing a scalable, automated paradigm for continuous self-improvement by generating high-quality and diverse training data.

    \item A key contribution is the scalable expansion of agent capabilities on complex, long-horizon tasks, which is enabled by our hindsight reproduction mechanism that learns efficiently from failures.
\end{itemize}

\section{Related Works}
\label{sec:relatedworks}

\textbf{GUI Agent Development and Limitations.} GUI agents are intelligent systems that interact with GUIs to perform automated tasks~\citep{wang2025guiagentsfoundationmodels}. Despite advances in intelligent cross-platform operations~\citep{nguyen2024guiagentssurvey}, GUI agent development~\citep{showui,openwebvoyager,cogagent,guithinker,2025gui,mobilea3gent} still depends on heavy human annotation~\citep{sun2025guixploreempoweringgeneralizablegui}, a key bottleneck. Recent works~\citep{agentq,wang2025sotalessmctsguidedsample} integrate MCTS and use both success and failure data via RL methods including DPO~\citep{dpo}, RFT~\citep{rft}, and PPO~\citep{ppo}. Yet these methods have flaws: trajectory-level DPO~\citep{rafailov2024directpreferenceoptimizationlanguage} and RFT underutilize search tree information. Step-level DPO~\cite{step-dpo} ignores intermediate step values by labeling them only as negatives. This restricts path exploration and produces rigid agents. In contrast, our work maximizes MCTS via hindsight reproduction, using all reasonable trajectories to guarantee high data quality and diversity.

\noindent\textbf{Reward Function.} Reward Models are crucial for aligning LLMs with human preferences and optimizing Virtual Agents' reasoning. However, Outcome Reward Models (ORMs)~\citep{yu2024ovmoutcomesupervisedvaluemodels,zhu2024starlingb,liu2025skyworkrewardv2scalingpreferencedata} often depend on costly manual annotation and solely judge final results, consequently lacking a detailed understanding of the intermediate process. This limitation has prompted the development of various process reward methodologies (PRMs)~\citep{qvalue,wang2024mathshepherdverifyreinforcellms,lightman2023letsverifystepstep}. For instance, AgentQ~\citep{agentq} utilizes self-prompting to select high-scoring actions but faces significant agent subjectivity, while ProgRM~\citep{zhang2025progrmbuildbettergui} employs progress increments as rewards, yet lacks multi-dimensional consideration. Recent works~\citep{bu2025limitsvirtualagentapplication,miao2025boostingvirtualagentlearning} have improved these issues by incorporating multi-dimensional evaluation metrics. Nevertheless, our work innovatively proposes a stepwise cycle-consistent reward within dual learning framework, which uniquely focuses on providing step-level action-intent aligned rewards to enhance the agent's internal comprehension of low-level actions, thereby synergistically boosting its capabilities.

\section{Method}


This section presents the design of \textit{Iron}, a model-agnostic Intent-aligned and Retrospective framework that empowers MLLMs to generate high-quality diverse planning trajectories and iteratively improve via self-training dual learning, as shown in Figure~\ref{fig:overview1_full_page}. We first introduce the dual learning task formulation in Section~\ref{sec:pre}, building two core co-enhanced agent capabilities. In Section~\ref{sec:3.2}, we elaborate on stepwise fine-grained alignment between low-level actions and underlying intents. Finally, in Section~\ref{sec:3.3}, we show how hindsight reproduction converts failed data into usable samples. Dual learning works closely with the hindsight reproduction mechanism, and the two components form a complete closed loop for autonomous learning and continuous evolution of the model.

\subsection{Preliminary and Task Formulation}\label{sec:pre}
Our approach aims to enhance performance of a generalist agent through two complementary task settings. We first demonstrate the modeling process and then define core tasks. In a high-level planning task, the agent proposes executable low-level actions from instructions, modeled as a Partially Observable Markov Decision Process (POMDP) defined by the quintuple $(\mathcal{S}, \mathcal{O}, \mathcal{A}, \mathcal{T}, \mathcal{R})$. Here, $\mathcal{S}$ represents the state space encompassing all possible states across platforms, $\mathcal{O}$ denotes the observation space consisting of textual or visual representations from the current device, and $\mathcal{A}$ includes all feasible actions. Central to this process is the transition function $\mathcal{T}(s_{t+1}|s_t, a_t)$, which describes the dynamics of moving from one state $s_t \in \mathcal{S}$ to another state $s_{t+1} \in \mathcal{S}$ after executing action $a_t$. The reward function $\mathcal{R}(s_t, a_t)$ provides feedback, guiding the agent towards optimal decisions.

Based on the aforementioned modeling process, we define two tasks for dual learning:


1) \textbf{\textit{Instruction Grounding:}} Given an instruction $I$, the observation $o_t$ at the current step $t$, and the sequence of previously executed actions $\tau_{t-1} = \{a_1, \ldots, a_{t-1}\}$, determine the next low-level action $a_{t}$ to execute. We formulate this as $\mathcal{P}(a_{t}|I, o_t, \tau_{t-1})$.

2) \textbf{\textit{Intent Understanding:}} Given a sequence of low-level actions $\tau_{t} = \{a_1, \ldots, a_{t-1}, a_{t}\}$ and the observation $o_t$ at the current step $t$, understand the underlying intent $U_t$ of the sequence. This is formulated as $\mathcal{D}(U_t|o_t, \tau_{t})$.

These two tasks are synergistic, designed to enhance an agent's decision-making capabilities in complex environments. At their core is a \textbf{virtuous cycle}: On one hand, improved instruction grounding yields more accurate action sequences, which helps the agent learn the logic between actions and high-level intent. On the other hand, robust intent extraction from these accurate sequences enhances the agent's comprehension of instructions and context, thereby directly reinforcing the grounding process.

\subsection{Stepwise Intent-aligned Dual learning}\label{sec:3.2}


This section details the full pipeline of the stepwise intent-aligned dual learning framework. This framework sequentially accomplishes model cold-start via imitation learning, performs high-quality trajectory exploration with fine-grained intent-action alignment through MCTS integrated with the SCC reward, and achieves iterative model enhancement via self-training dual learning, constructing a core synergistic cycle for bidirectional reinforcement of instruction execution and intent understanding. This pipeline works closely with the hindsight reproduction mechanism in Section \ref{sec:3.3}, and the two components form a complete closed loop for autonomous learning and continuous evolution of the model.

\noindent\textbf{Imitation Learning.} We adopt Imitation Learning (IL) as a critical cold-start stage to bootstrap the backbone model prior to the launch of the dual learning framework. Though the backbone model has inherent grounding capabilities, it lacks training on task planning and intent-action alignment data, a prerequisite for the agent's data collection. We leverage expert demonstration trajectories $\tau$ to train the agent on the two core capabilities defined in Section~\ref{sec:pre}: \textit{Instruction Grounding} and \textit{Intent Understanding}. Given the task instruction $I$ and current observation $o_t$, the IL objective is to maximize the log-likelihood of generating the correct action $a_t$ and its corresponding intent $U_t$ at each time step $t$, formulated as:
\begin{equation*}
\begin{aligned}
\mathcal{J}_{IL}(\theta)=\mathbb{E}_{(I, \tau) \sim D_{IL}} \sum_{t=1}^{T}\big[&\log \pi_{\theta}(a_{t} \mid I, o_t, U_{t})\\
&+\log \pi_{\theta}(U_{t} \mid I, o_t)\big].
\end{aligned}
\end{equation*}
where $\theta$ denotes the backbone model parameters, $D_{IL}$ is the expert trajectory dataset, and $T$ is the total length of the planning trajectory. More training details and prompts can be found in Appendix A.

\noindent\textbf{MCTS Data Collection.}
The MCTS algorithm employed here largely follows the design of~\citet{RAP}, consisting of the standard four phases: selection, expansion, simulation, and backpropagation. Given MCTS's inherent superiority in balancing exploration and exploitation, it systematically investigates various actions and assesses their outcomes through numerous simulations. This disciplined approach makes MCTS highly suitable for acquiring a diverse range of high-quality trajectories by consistently selecting the optimal action for execution. However, MCTS suffers from sparse reward signals, where simple outcome-based rewards are ineffective for refining stepwise actions. While many methods attempt to enhance single-step supervision by constructing a reward model, these models often introduce new dependencies by relying on closed-source systems, externally trained models, or human judgments to generate rewards.

\noindent\textbf{Stepwise Cycle-Consistent (SCC) reward mechanism.} To address this limitation, and guided by the cycle-consistent principle~\cite{cyclegan} that the inferred user intent $U$ must be semantically equivalent to the original instruction $I$, we propose the SCC reward mechanism without relying on any human intervention or external resources. Specifically, given a user instruction $I$, we start from an initial state $s_0$ with the goal of reaching a final state $s^*$. Each state $s_t$ includes the history of past actions $\tau_{t-1} = \{a_1, \ldots, a_{t-1}\}$ and current observation $o_t$. At step $t$, the agent expands $n$ child node actions $\{a_{t}^{i} \mid i=1..n\}$ where $a_{t}^{i} \sim \mathcal{P}(I, {\tau_{t-1}}, o_t)$ for $i=1$ to $n$ as candidate choices. For each $a_{t}^{i} \in$ the expanded action set $A_t$, based on cycle-consistency, we use the matching score between the inferred intent $U_{t}^{i}$ and instruction $I$ as the child node reward; a higher score indicates a greater success likelihood of the action sequence, guiding the agent's selection at step $t$. Let $z_{\text{align}}$ and $z_{\text{misalign}}$ denote the raw logits of semantic alignment and misalignment output by the agent's self-prompting mechanism, the semantic alignment probability is:
$$
p\left(\text{align} \mid I, \mathcal{D}(o_t, \tau_{t-1}, a_t^i)\right) = \frac{\exp(z_{\text{align}})}{\exp(z_{\text{align}})+\exp(z_{\text{misalign}})}.
$$
The SCC reward is formalized as:
$$
\begin{aligned}
U_{t}^{i} &= \mathcal{D}(o_t, \tau_{t-1}, a_t^i), \\
\mathcal{SCC}(s_t, a_t^i) &= \sigma\left( \log \frac{p(\text{align} \mid \cdot)}{1-p(\text{align} \mid \cdot)} \right) = \sigma\left( z_{\text{align}} - z_{\text{misalign}} \right),
\end{aligned}
$$
where $\mathcal{D}(\cdot)$ denotes the step-wise intent inference function defined in Section~\ref{sec:pre}, and $\sigma(\cdot)$ is the Sigmoid function that normalizes the final semantic reward value to the standard range $[0,1]$. The complete workflow of our proposed SCC reward mechanism is illustrated in Figure~\ref{fig:overview1_full_page}b, and detailed implementation procedures and prompt templates are provided in Appendix C.

This reward mechanism tends to select the action with the highest matching score for expansion, providing supervision for each trajectory segment's action-intent alignment, which in turn encourages the agent to generate more logical and reasonable actions. The algorithm and parameters for MCTS integration are detailed in Appendix B, and we will present experiments on the reliability of the SCC reward in subsequent sections.

\noindent\textbf{Self-training Dual Learning.} The core of self-training dual learning lies in leveraging the bidirectional synergy between the two pre-defined complementary tasks, integrating cycle consistency constraints to realize iterative mutual enhancement while embedding the entire process into a self-training loop that continuously refines data quality. Drawing on the fundamental paradigm of dual learning, we establish a closed-loop optimization mechanism where the two tasks not only learn from each other through mutual supervision but also jointly benefit from the high-quality data generated by iterative self-training, addressing the limitations of isolated task optimization and static dataset dependence.

We formalize the dual learning process by defining a joint objective function that unifies the optimization of both tasks and incorporates cycle consistency constraints consistent with the SCC reward mechanism. Let $\theta$ denote the shared parameters of the backbone model, $\mathcal{D}_{ST}$ represent the self-training dataset composed of refined trajectories from each iteration, $I$ be the task instruction, $o_t$ the current observation, $\tau_{t-1}$ the action history up to step $t-1$, $a_t^*$ the expert action, and $U_t^*$ the ground-truth intent. The joint objective function is formulated as:
\begin{equation*}
\begin{aligned}
\mathcal{J}_{DL}(\theta) &= \mathbb{E}_{(I, o_t, \tau_{t-1}, a_t^*, U_t^*) \sim \mathcal{D}_{ST}} \Bigg[ 
    \log \pi_{\theta}(a_t^* | I, o_t, \tau_{t-1}) \\
    &\quad + \log \pi_{\theta}(U_t^* | o_t, \tau_{t-1}) + \lambda \cdot \mathcal{L}_{cyc}(I, U_t) 
\Bigg]
\end{aligned}
\end{equation*}
where the first two terms maximize the log-likelihood of \textit{Instruction Grounding} and \textit{Intent Understanding} respectively. The third term $\mathcal{L}_{cyc}(I, U_t)$ is the cycle consistency loss, which quantifies the semantic discrepancy between the instruction $I$ and inferred intent $U_t$ (consistent with the alignment metric in SCC reward), with $\lambda$ as the weight coefficient.

The iterative self-training process optimizes dual learning via a recursive cycle of data generation, refinement, and model update. In each iteration, the current model parameterized by $\theta_{k-1}$ interacts with the environment via MCTS to generate raw trajectories, where the SCC reward—derived from the cycle consistency loss and a core framework component—guides action selection and trajectory quality assessment. These raw trajectories are processed through success selection, hindsight reproduction of failed trajectories, and invalid filtering with five metrics to form the high-quality self-training dataset $\mathcal{D}_{ST}^{(k)}$. This dataset merges with the cumulative dataset from prior iterations $\mathcal{D}_{ST}^{(\leq k-1)}$ to form the complete training data $\mathcal{D}_{ST}^{(\leq k)}$, on which the model optimizes the joint objective $\mathcal{J}_{DL}(\theta)$ to obtain the updated parameter $\theta_k$.

This integrated framework delivers two key advantages that drive its superior performance. The cycle consistency constraint inherent to dual learning ensures fine-grained alignment between low-level actions and high-level intents. This mitigates the model’s mechanical imitation behavior and enhances its ability to generalize to unseen scenarios by capturing the underlying logic linking instructions, actions, and intents. The iterative self-training loop continuously supplements dual learning with high-quality and diverse data. Successful trajectories provide reliable positive supervision while repurposed failed trajectories expand task coverage and data diversity. This creates a clear bootstrap effect where improved models generate better data that in turn further refines the model. Experimental results verify that this synergy enables consistent performance gains across multiple benchmarks.

\begin{figure}[t!] 
    \centering     
    \includegraphics[width=\columnwidth]{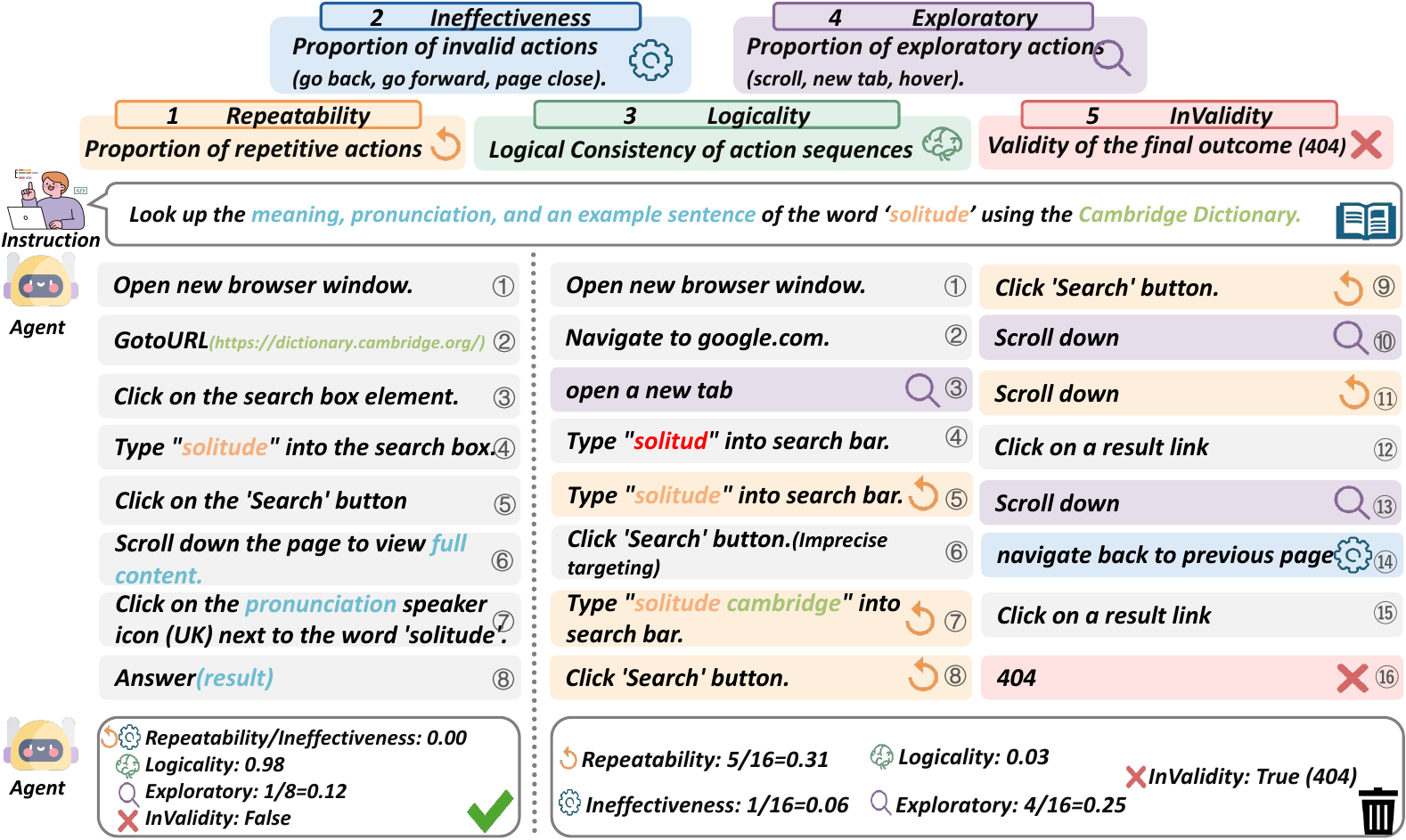} 
    
    \caption{\textbf{An example illustrating the positive (left) and negative (right) filtering of trajectories using five meticulously designed metrics.} By filtering either already collected successful or failed trajectories (after modifying their objectives), we ensure dataset quality and achieve further improvements in accuracy.} 
    \label{fig:filter} 
\end{figure}

\subsection{Hindsight Reproduction}\label{sec:3.3}


This section mainly addresses the efficient utilization and quality control of exploration trajectories during the dual learning process, and designs standardized processing strategies for both successful and failed trajectories generated by MCTS. We define the screening criteria for successful trajectories, convert failed trajectories through hindsight reproduction, and filter invalid samples with multi-dimensional metrics to finally generate high-quality and diverse self-training data. The optimized data is fed back into the dual learning framework to support iterative model optimization, thereby completing the full closed loop of self-supervised learning.

\noindent\textbf{Successful Trajectory Inclusion.} For \textit{Instruction Grounding}, our SCC reward mechanism provides supervision and reward signals by calculating the alignment between the original instruction \(I\) and the understood intent \(U\) for action sequences. When this reward signal exceeds a predefined threshold \(\delta\), the trajectory is considered to be added to the training set. This condition can be formalized as:
$$
    \text{\textit{Match}}(I,\mathcal{D}(o_t,a_1, \ldots, a_{t-1}, a_{t}^{i}))>\delta
$$


For \textit{Intent Understanding}, we employ a self-reflective prompting at state \(s_{t+1}^i\), which results from executing the action \(a_t^i\), to provide reference for understanding the low-level sequences $\tau_{{t}}^{i}=\{a_1, \ldots, a_{t-1}\ ,a_{t}^{i}\}$.  Specifically, we utilize a self-prompting mechanism to generate the ground truth, and its response serves as the ground truth for the intent understanding of the action sequence up to time \(t\). Due to space constraints, the specific prompt content is detailed in Appendix D.1.

\noindent\textbf{Failure Hindsight Reproduction.} To fully leverage the abundant failed data generated during the search process, we adopt a hindsight reproduction strategy within MCTS to transform failed trajectories into successful data, as illustrated in Figure~\ref{fig:overview1_full_page}c. Additionally, we ensure the quality of the training data by introducing five metrics to filter out invalid trajectories from the dataset. There is an example shown in the Figure~\ref{fig:filter}.

A crucial observation for our method is that no trajectory is truly a failure. Failure is simply relative to the intended objective. Even when a trajectory exhausts all attempts at its original goal, the sequence of actions taken \textbf{inherently results in a valid, concrete outcome.} This allows us to repurpose these perceived failures. Instead of discarding the data, we relabel the trajectory, treating this actual outcome as its new, successfully achieved goal. Leveraging the two complementary tasks within the dual-learning framework, we further refine this process. By implementing an adaptation of the HER~\cite{her} mechanism, we treat the intent and description ($U$) derived from the entire failed trajectory as its new goal. As illustrated in Figure~\ref{fig:overview1_full_page}c, we take the expansion of two child nodes in MCTS as an example. When the search tree reaches its maximum depth without task completion, we then apply hindsight reproduction to retrace the path and convert the failed trajectory into a successful one with a relabeled goal. These newly successful trajectories are then filtered and added to the training set. This approach not only salvages valid data from failures but also expands the diversity of high-quality training examples. The process can be modeled as:
$$
I_{new}=U_{end}=\mathcal{D}(o_{end}, a_1, \ldots, a_{end-1}, a_{end})
$$

\noindent\textbf{Trajectory Filtering.} Due to the complexity of the digital environment, there may be numerous unpredictable feedback loops causing the model to repeatedly perform certain actions, or external factors such as network issues, leading to invalid trajectories. To address this, we introduce five metrics to filter out such problematic trajectories:
1) \textit{Repeatability}: proportion of semantically and behaviorally similar repetitive actions. 2) \textit{Ineffectiveness}: proportion of ineffective actions such as \texttt{go back} or \texttt{page close}. 3) \textit{Exploratory}: proportion of exploratory actions such as \texttt{scroll} or \texttt{hover}. 4) \textit{Logicality}: logical consistency of an action sequence compared to human cognitive patterns. 5) \textit{Invalidity}: whether the final outcome is invalid due to errors like 404 or network issues. Figure~\ref{fig:filter} showcases the efficacy of these metrics by providing a clear illustrative example, contrasting valid and invalid trajectories after the filtering process. The detailed calculation of the five metrics can be found in Appendix D.2.

\begin{table*}[h!]
\centering

\renewcommand{\arraystretch}{0.9}

\resizebox{\textwidth}{!}{%
\tiny
\begin{tabular}{l l *{6}{c} | c}
\toprule

\textbf{Category} & \textbf{Agent} & 
\multicolumn{6}{c|}{\textbf{OSWorld}} & 
\textbf{AndroidWorld} \\

\cmidrule(lr){3-8} \cmidrule(lr){9-9}

& & OS & Office & Daily & Profess & Workflow & Avg. & Avg. \\

\midrule


\multirow{8}{*}{\textbf{Closed-source}}

& GPT-4o 
& 8.33\% & 3.58\% & 6.07\% & 4.08\% & 5.58\% & 5.03\% & 29.52\% \\

& GPT-4V 
& 12.50\% & 1.86\% & 7.58\% & 4.08\% & 6.04\% & 5.26\% & 30.47\% \\

& Gemini-ProV 
& 8.33\% & 3.58\% & 6.55\% & 16.33\% & 2.08\% & 5.80\% & 21.40\% \\

& Gemini-Pro-1.5 
& 12.50\% & 6.99\% & 2.71\% & 6.12\% & 3.60\% & 5.40\% & 20.80\% \\

& Claude-3-Opus 
& 4.17\% & 1.87\% & 2.71\% & 2.04\% & 2.61\% & 2.42\% & 24.70\% \\

& Claude-3-7-Sonnet
& 50.00\% & 35.87\% & 46.10\% & 46.94\% & 17.66\% & 35.83\% & 55.82\% \\

& Claude-4-Sonnet
& 45.83\% & 43.56\% & 55.13\% & 55.10\% & 28.49\% & 43.88\% & 63.46\% \\

& Claude-Sonnet-4-5
& 70.83\% & 62.41\% & 57.65\% & 63.27\% & 46.99\% & 58.08\% & 68.91\% \\

\midrule


\multirow{18}{*}{\textbf{Open-source}}

& Qwen2.5-VL-72B
& 8.33\% & 2.56\% & 7.69\% & 0.00\% & 5.37\%
& 4.43\% & 31.20\% \\

& Qwen2.5-VL-32B
& 8.33\% & 1.71\% & 7.69\% & 2.04\% & 0.00\%
& 3.04\% & 23.90\% \\

\cmidrule(lr){2-9}
& Qwen3-VL-8B-IL
& 62.50\% & 31.58\% & 45.71\% & 58.70\% & 24.75\% 
& 38.39\% & 33.47\% \\

& Qwen3-VL-8B-\textit{Iron-10k}
& 66.67\% & 32.63\% & 47.14\% & 60.87\% & 25.74\%
& 39.28\% & 35.85\% \\

& Qwen3-VL-8B-\textit{Iron-20k}
& 70.83\% & 34.74\% & 50.00\% & 63.04\% & 26.73\%
& 40.47\% & 37.40\% \\

& Qwen3-VL-8B-\textit{Iron-30k}
& \textbf{75.00}\% & 36.84\% & \textbf{52.86}\% & \textbf{67.39}\% & 28.71\%
& \textbf{41.96}\% & \textbf{39.15}\% \\

& Qwen3-VL-8B-\textit{Full}
& 50.00\% & \textbf{42.11}\% & 44.29\% & 43.48\% & 36.63\%
& 41.67\% & 37.88\% \\

& Qwen3-VL-8B-\textit{DPO}
& 41.67\% & 38.95\% & 41.43\% & 41.30\% & \textbf{39.60}\%
& 40.18\% & 35.69\% \\

\cmidrule(lr){2-9}

& OS-Atlas-7B-IL
& 8.33\% & 2.10\% & 4.29\% & 2.17\% & 1.98\%
& 2.97\% & 17.14\% \\

& OS-Atlas-7B-\textit{Iron-10k}
& 8.33\% & 3.15\% & 5.71\% & 2.17\% & 1.98\%
& 3.57\% & 18.09\% \\

& OS-Atlas-7B-\textit{Iron-20k}
& \textbf{12.50}\% & 3.15\% & 5.71\% & 6.52\% & 2.97\%
& 4.76\% & 19.51\% \\

& OS-Atlas-7B-\textit{Iron-30k}
& \textbf{12.50}\% & \textbf{4.21}\% & \textbf{7.14}\%
& \textbf{8.69}\% & 3.96\%
& \textbf{5.95}\% & \textbf{21.13}\% \\

& OS-Atlas-7B-\textit{Full}
& \textbf{12.50}\% & 3.15\% & \textbf{7.14}\%
& 6.52\% & \textbf{4.95}\%
& 5.65\% & 19.06\% \\

& OS-Atlas-7B-\textit{DPO}
& 8.33\% & 4.21\% & 5.71\%
& 6.52\% & 2.97\%
& 4.16\% & 17.95\% \\

\cmidrule(lr){2-9}

& InternVL2.5-4B-IL
& 4.16\% & 1.05\% & 2.85\% & 0.00\% & 0.99\%
& 1.49\% & 8.71\% \\

& InternVL2.5-4B-\textit{Iron-10k}
& 4.16\% & 1.05\% & 4.29\% & 2.17\% & 0.99\%
& 2.08\% & 9.29\% \\

& InternVL2.5-4B-\textit{Iron-20k}
& 4.16\% & 2.10\% & 4.29\% & 2.17\% & 1.98\%
& 2.67\% & 10.12\% \\

& InternVL2.5-4B-\textit{Iron-30k}
& \textbf{8.33}\% & \textbf{3.15}\% & \textbf{5.71}\%
& \textbf{4.37}\% & 1.98\%
& \textbf{3.87}\% & \textbf{11.66}\% \\

& InternVL2.5-4B-\textit{Full}
& \textbf{8.33}\% & 2.10\% & 4.29\%
& 2.17\% & \textbf{2.97}\%
& 3.37\% & 11.64\% \\

& InternVL2.5-4B-\textit{DPO}
& 4.16\% & 2.10\% & 2.85\%
& 2.17\% & 1.98\%
& 2.38\% & 9.72\% \\

\bottomrule

\end{tabular}%
}
\caption{
\textbf{Model Performance on OSWorld and AndroidWorld.} \textbf{Bold} values indicate the best performance among open-weight models.
IL denotes imitation learning baselines. \textit{Iron} variants use 10k, 20k, and 30k self-training samples.
\textit{Full} uses 3$\times$ IL data for SFT, and \textit{DPO} uses Step-DPO.
Detailed baseline settings and statistical values are provided in Appendix F.2 and F.3.
}
\label{tab:model_performance_custom}
\end{table*}
\section{Experiment}

\textit{Iron} is a model-agnostic method that synergistically enhances GUI agents' understanding and planning capabilities through self-training and dual learning. Our approach yields a generalist agent that achieves improved performance across desktop, web, and mobile environments simultaneously. Our evaluation comprises five main sections: first, we detail the overall experimental setup in Section~\ref{sec:4.1}; then we comparatively analyze the agent's planning capabilities against various planning baselines in Section~\ref{sec:4.2}; subsequently, we assess the agent's generalization in unseen environment in Section~\ref{sec:4.3}; we provide a case study in Section~\ref{sec:4.5}; finally, we analyze the reliability of key components in Section~\ref{sec:reliability_scc}.

\begin{table}[t!]
\centering

\renewcommand{\arraystretch}{1.02} 

\resizebox{\linewidth}{!}{
\begin{tabular}{l c c c c}
\toprule
\textbf{Agent} & \multicolumn{4}{c}{\textbf{VisualWebArena}} \\
\cmidrule(lr){2-5}
& Classifieds & Reddit & Shopping & Overall \\
\midrule
GPT-4o & 11.31\% & 13.67\% & 14.33\% & 13.25\% \\
GPT-4V & 10.26\% & 13.88\% & 12.16\% & 12.54\% \\
\midrule

Qwen3-VL-8B-IL 
& 12.45\% & 14.86\% & 15.62\% & 14.31\% \\

Qwen3-VL-8B-\textit{Iron-10k}
& 13.28\% & 15.73\% & 16.44\% & 15.15\% \\

Qwen3-VL-8B-\textit{Iron-20k}
& 14.16\% & 16.82\% & 17.53\% & 16.25\% \\

Qwen3-VL-8B-\textit{Iron-30k}
& \textbf{15.02\%} & \textbf{18.01\%} & \textbf{18.72\%} & \textbf{17.38\%} \\

Qwen3-VL-8B-\textit{Full}
& 14.73\% & 17.56\% & 18.26\% & 16.85\% \\

\midrule
OS-Atlas-7B-IL & 6.86\% & 8.17\% & 8.38\% & 7.94\% \\
OS-Atlas-7B-\textit{Iron-10k} & 7.29\% & 8.65\% & 8.60\% & 8.27\% \\
OS-Atlas-7B-\textit{Iron-20k} & 8.07\% & 9.61\% & 9.24\% & 9.05\% \\
OS-Atlas-7B-\textit{Iron-30k} & \textbf{8.58\%} & 10.09\% & \textbf{10.53\%} & 9.93\% \\

OS-Atlas-7B-\textit{Full} & 8.15\% & \textbf{10.57\%} & 10.96\% & \textbf{10.26\%} \\
\midrule
InternVL2.5-4B-IL & 3.43\% & 4.32\% & 4.30\% & 4.08\% \\
InternVL2.5-4B-\textit{Iron-10k} & 3.86\% & 4.81\% & 4.94\% & 4.63\% \\
InternVL2.5-4B-\textit{Iron-20k} & 4.29\% & 5.77\% & 5.37\% & 5.18\% \\
InternVL2.5-4B-\textit{Iron-30k} & \textbf{4.72\%} & \textbf{6.25\%} & 6.23\% & \textbf{5.96\%} \\

InternVL2.5-4B-\textit{Full} & 4.29\% & \textbf{6.25\%} & \textbf{6.66\%} & 5.84\% \\
\bottomrule
\end{tabular}
} 
\caption{
\textbf{Model Performance Comparison on VisualWebArena.} Due to space constraints, the model configurations in the table follow those in Section~\ref{sec:4.2}.
}
\label{tab:model_performance_visualwebarena}
\end{table}

\subsection{Experimental Setup}\label{sec:4.1}
\textbf{Model setup.} We employed the Qwen3-VL-8B~\citep{qwen3vl}, OS-Atlas-Base-7B~\citep{osatlas} and InternVL2.5-4B~\citep{internvl25} models as backbones. These models, representing varying sizes and architectures, were used to demonstrate that \textit{Iron} is a model-agnostic framework capable of achieving self-improvement through self-training.

\noindent\textbf{Training set.} Our training methodology involves two distinct stages: Imitation Learning and Self-Training.

1) \textbf{Imitation Learning Stage:} Although our model possessed grounding capabilities, it was not trained on planning data. Since our framework requires the agent to navigate MCTS to achieve goals, we provided planning ability by integrating 120k grounding data with 90k planning data from the Aguvis~\citep{aguvis} dataset (stage2). Training details and prompts can be found in Appendix A.

 2) \textbf{Self-Training Stage:} Following the IL stage, we progressively refined the agent through iterative training, using 30k high-quality data through gradual exploration, thereby enhancing planning and understanding abilities.

\noindent\textbf{Evaluation Benchmarks.} We evaluate our generalist agent on three cross-platform benchmarks across diverse, complex tasks: OSWorld~\citep{osworld} for desktop and web tasks, AndroidWorld~\citep{android_world} for mobile tasks, and VisualWebArena~\citep{koh2024visualwebarena} for generalization to unseen web environments. In all evaluations, the agent perceives and interacts solely through raw screenshots.

\noindent\textbf{Baseline Selection.} 
Different from task-specific methods that aim to achieve the best performance with a particular backbone, \textit{Iron} is designed as a \textbf{model-agnostic self-training framework} that can integrate with various foundation models. Therefore, our experiments focus on validating whether \textit{Iron} consistently improves different base agents through autonomous data generation and iterative self-training, rather than optimizing for a single strongest backbone. For controlled comparisons, variants within each backbone share the initialization, evaluation protocol, and IL data scale. \textit{Full} uses $3\times$ IL data for supervised fine-tuning as a data-scaling baseline, while \textit{DPO} leverages preference pairs naturally generated during trajectory collection with SCC guidance.
\vspace{-0.2cm}
\subsection{Evaluation Across Planning Benchmarks}
\label{sec:4.2}

We report the agents' performance in Table~\ref{tab:model_performance_custom} and summarize the main findings below. \noindent\textbf{1) Competitive Performance with Limited Annotations.}
Models enhanced by \textit{Iron} consistently outperform the $\textit{Full}$ versions trained on three times more SFT data, showing that self-exploration and hindsight reproduction produce informative data of higher quality than simply scaling passive annotations. On the stronger Qwen3-VL-8B backbone, \textit{Iron-30k} achieves 41.96\% on OSWorld and 39.15\% on AndroidWorld, surpassing the $\textit{Full}$ model with fewer training samples. Our pipeline constructs preference pairs using SCC reward scores. Under the same backbone and hyperparameters, \textit{Iron} also outperforms Step-DPO trained on 30k pairs, indicating that action--intent cycle consistency provides a finer-grained and reliable learning signal.

\noindent\textbf{2) Consistent Improvement via Iterative Self-Training.}
Performance improves as \textit{Iron} data increases across different model backbones. On Qwen3-VL-8B, the OSWorld score increases from 38.39\% to 41.96\%, while AndroidWorld improves from 33.47\% to 39.15\% with expanded self-training data. Similar trends are observed on OS-Atlas-7B and InternVL2.5-4B, demonstrating the model-agnostic effectiveness of our framework. Table~\ref{tab:iterative_vs_non_iterative} further shows that iterative self-training outperforms one-shot SFT using the same data. The widening gap from 20k to 30k samples suggests that one-shot SFT reaches a performance ceiling, whereas iterative refinement benefits from larger and cleaner datasets.
\begin{table}[t!]
    \centering
   
    \resizebox{\columnwidth}{!}{%
        \footnotesize 
        \setlength{\tabcolsep}{3pt} 
        \begin{tabular}{@{}ll|c|c|c@{}} 
            \toprule
            \textbf{Agent} & \textbf{Method} & \textbf{OSworld} & \textbf{AndroidWorld} & \textbf{VWA} \\
            \midrule
            OS-Atlas-IL & Baseline & 2.97\% & 17.14\% & 7.94\% \\
            \midrule
            OS-Atlas-20k & Non-Iterative & 4.46\% & 19.39\% & 8.83\% \\
            OS-Atlas-\textit{Iron}\textit{-20k} & Iterative & \textbf{4.76\%} & \textbf{19.51\%} & \textbf{9.05\%} \\
            \midrule 
            OS-Atlas-30k & Non-Iterative & 5.03\% & 19.49\% & 9.01\% \\
            OS-Atlas-\textit{Iron}\textit{-30k} & Iterative & \textbf{5.95\%} & \textbf{21.13\%} & \textbf{9.93\%} \\
            \bottomrule
        \end{tabular}%
    } 
     \caption{
        \textbf{Performance Comparison of Iterative and Non-Iterative Self-Training.} Models labeled with \textit{Iron} employ our continuous iterative training procedure, whereas other models are trained in a non-iterative, one-shot manner.
    }
    \label{tab:iterative_vs_non_iterative}
    
\end{table}

\noindent\textbf{3) Enhanced Performance on Hard Tasks.}
On OSWorld, \textit{Iron}'s advantages are pronounced in challenging ``Professional'' (e.g., VS Code and GIMP) and ``Office'' tasks (e.g., LibreOffice). These tasks require long-horizon planning and precise multi-step interactions, where \textit{Iron}-trained models achieve substantial gains, with metrics surpassing GPT-4 series models. This demonstrates that our self-training mechanism improves the robustness and generalization needed for complex tasks in professional software environments.

\subsection{Generalization in Unseen Environment}\label{sec:4.3}
Table~\ref{tab:model_performance_visualwebarena} summarizes the experimental results on unseen environments, demonstrating that self-training and hindsight enhance generalization. The \textit{Iron} framework shows robust transferability to the unseen VisualWebArena benchmark. On Qwen3-VL-8B, \textit{Iron-30k} improves the overall success rate from 14.31\% to 17.38\%, outperforming the \textit{Full} model trained with more SFT data. Similar improvements are observed on OS-Atlas-7B and InternVL2.5-4B, with gains of 25\% and from 4.08\% to 5.96\%, respectively. This efficacy is primarily attributed to our novel Hindsight Reproduction mechanism. Rather than solely learning from successes, \textit{Iron} repurposes failed trajectories by reformulating them into novel, successful tasks. This process generates a diverse training curriculum that enables the agent to master broader strategies and intent combinations.

\begin{figure}[t!]
    \centering

    \begin{subfigure}[b]{0.25\columnwidth}
        \centering
        \includegraphics[
            width=\linewidth,
            height=4.5cm,
            keepaspectratio
        ]{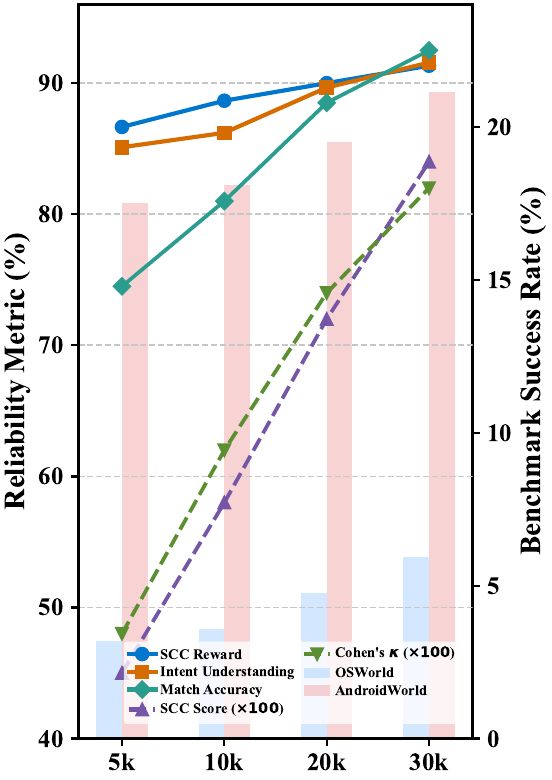}
        \caption{}
        \label{fig:two_metrics}
    \end{subfigure}%
    \hspace{-0.02\columnwidth}%
    \begin{subfigure}[b]{0.75\columnwidth}
        \centering
        \includegraphics[
            width=\linewidth,
            height=4.5cm,
            keepaspectratio
        ]{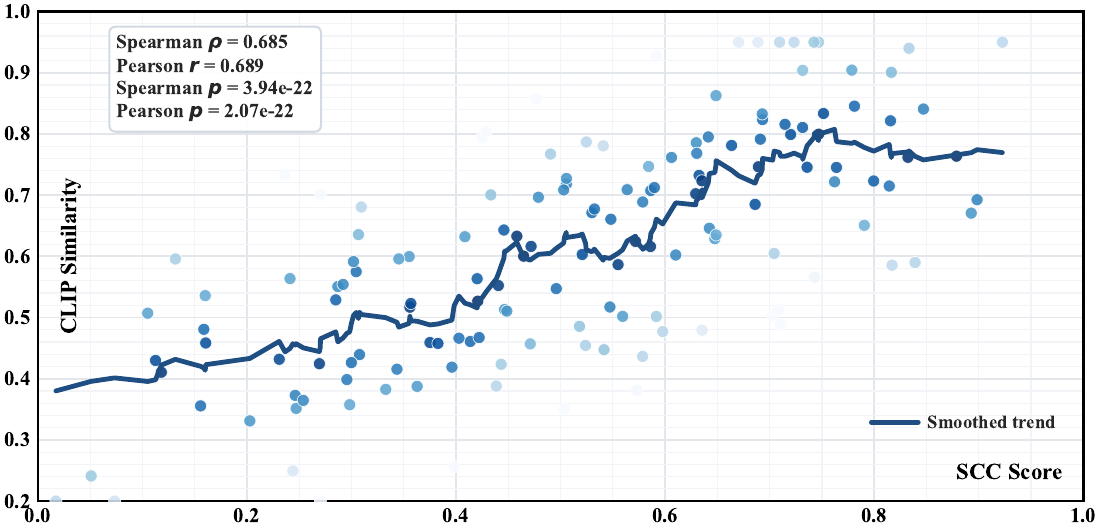}
        \caption{}
        \label{fig:scc_clip_similarity}
    \end{subfigure}

\caption{\textbf{Reliability analysis of SCC reward and intent understanding.}
    (a) Improvements in SCC reward, intent understanding, independent validation
    metrics, and downstream performance across self-training iterations.
    (b) Correlation between SCC scores and CLIP-based instruction--intent
    similarity, providing semantic validation of SCC reward.}
    \label{fig:module_validation}
    \vspace{-0.4cm}
\end{figure}
\subsection{Case Study: Enhancing Intent Understanding}\label{sec:4.5}

\begin{figure}[t!] 
    \centering     
    \includegraphics[width=\columnwidth]{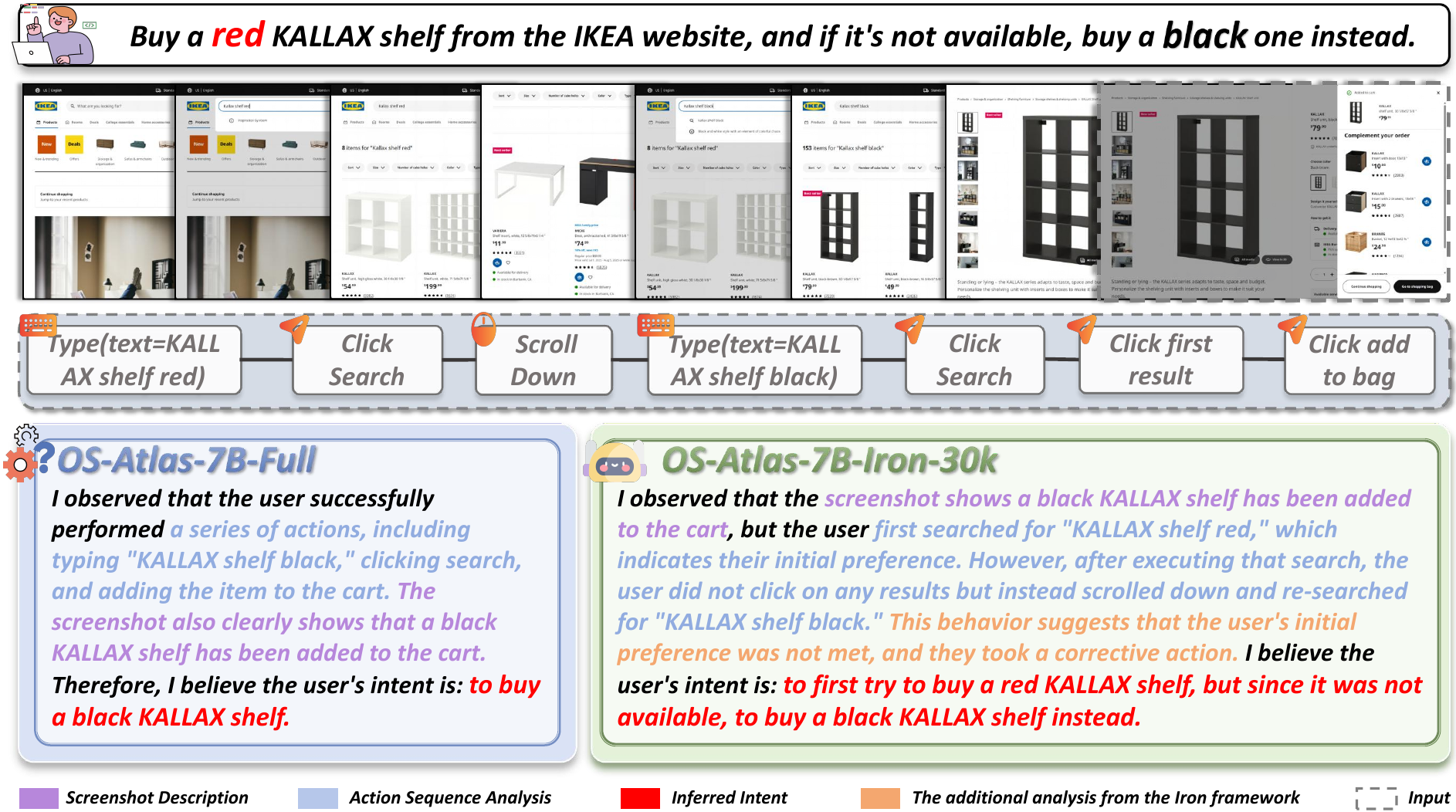} 
    
 \caption{A qualitative analysis showing superior intent understanding of our \textit{Iron}-trained model over baselines.} 
    \label{fig:cs} 
    \vspace{-0.3cm}
\end{figure}

To demonstrate the superior intent understanding of our \textit{Iron} framework for GUI agents, we conducted a case study shown in Figure~\ref{fig:cs}. We compared our self-trained model, OS-Atlas-7B-\textit{Iron-30k}, with the SFT traditional baseline. While the baseline model mechanically focuses on the final successful action, failing to grasp the user's initial preference, \textit{Iron} integrates the entire user journey. It synthesizes the initial failed search for the red shelf and the corrective search for the black one into a single, complete intent, proving its ability to understand complex, conditional goals.

\subsection{Reliability Analysis of SCC Reward and Intent Understanding}
\label{sec:reliability_scc}

The SCC reward and intent understanding module provide the core self-supervision
in \textit{Iron}. We evaluate their reliability across training iterations
using manual assessment and independent evaluators, as shown in
Figure~\ref{fig:two_metrics}.

\noindent\textbf{Training-stage reliability.}
Across 150 samples per iteration, SCC Match accuracy increases from 74.5\% at
5k steps to 92.5\% at 30k steps, while agreement with GPT-5.4 rises from
Cohen's $\kappa=0.48$ to $0.82$. Intent consistency on 80 OmniACT samples also
improves from 85.1\% to 91.6\%. Meanwhile, the SCC score grows from 0.45 to
0.84, accompanied by consistent gains on OSWorld and AndroidWorld. These trends
indicate that SCC tracks downstream competence rather than reinforcing the
model's own predictions.

\noindent\textbf{External semantic validation.}
Using CLIP ViT-B/32 as an independent evaluator, SCC scores correlate strongly
with instruction--intent similarity (Spearman $\rho=0.685$ and Pearson
$r=0.689$, both $p<0.001$; Figure~\ref{fig:scc_clip_similarity}). This
out-of-loop evidence confirms that SCC captures meaningful semantic alignment
and mitigates self-confirmation bias. Further SCC reliability analysis is provided in Appendix~C.3.

\section{Ablation Studies}\label{sec:5}
To quantify the contribution of each core component of the \textit{Iron} framework, we conducted ablation experiments. These removed key innovations. We validate filtering metrics and optimized thresholds in later analysis. All experiments used the OS-Atlas-7B-\textit{Iron-30k} model and were evaluated on OSWorld and AndroidWorld subsets, with results in Table~\ref{tab:ablation_results}.

\begin{table}[t!]
 \centering

 \renewcommand{\arraystretch}{1.2}

 \resizebox{\columnwidth}{!}{%
 \begin{tabular}{l c | c} 
 \toprule
 \textbf{Model} & \textbf{OSWorld} & \textbf{AndroidWorld}  \\
 \midrule
 
 a. \textbf{\textit{Iron}} & \textbf{5.95\%} & \textbf{21.13\%} \\ 
 b. w/o \textit{Dual Learning.} & 5.65\% ($\downarrow$ 5.0\%) & 20.01\% ($\downarrow$ 5.3\%) \\
 c. w/o \textit{SCC Reward.} & 5.35\% ($\downarrow$ 10.1\%) & 19.66\% ($\downarrow$ 7.0\%) \\
 d. w/o \textit{Hindsight Reprod.} & 5.06\% ($\downarrow$ 15.0\%) & 18.79\% ($\downarrow$ 11.1\%) \\
 e. w/o \textit{Trajectory Filter.} & 2.08\% ($\downarrow$ 65.0\%) & 10.94\% ($\downarrow$ 48.2\%) \\
 
 \bottomrule
 \end{tabular}%
 } 
  \caption{\textbf{Ablation study of \textbf{\textit{Iron}} components on OSWorld and AndroidWorld.} Parentheses show performance drop relative to the full model (row a).}
 \label{tab:ablation_results}
\end{table}

\noindent Our ablation study reveals the distinct contributions of \textit{Iron}'s core components. Removing the Dual Learning mechanism and SCC Reward leads to consistent performance degradation (5.0\% to 7.0\%) on both benchmarks, confirming their roles in enhancing action--intent alignment and providing dense optimization signals. Removing Hindsight Reproduction results in further performance drops, highlighting its importance in leveraging unsuccessful trajectories for improving exploration diversity. The largest degradation is caused by removing the Trajectory Filter. Specifically, without trajectory filtering, performance decreases by 65.0\% on OSWorld and 48.2\% on AndroidWorld, demonstrating that data quality is critical for stable self-improvement. These results confirm that the Trajectory Filter is essential for constructing reliable training signals in iterative self-training.


\section{Conclusion}
We propose \textbf{\textit{Iron}}, a novel model-agnostic self-training framework that, through intention-action alignment and hindsight reproduction, enables agents to learn autonomously from both successful and failed real-world experiences. As a highly scalable pipeline, \textit{Iron} achieved considerable performance gains on diverse cross-platform tasks and also made substantial progress in challenging unseen web scenarios. In summary, \textit{Iron} provides a powerful and scalable self-training method to autonomously enhance agents' core capabilities.

{
    \small
    \bibliographystyle{ieeenat_fullname}
    \bibliography{main}
}


\end{document}